\documentclass[runningheads]{llncs}
\usepackage[T1]{fontenc}
\usepackage{graphicx}%
\usepackage{multirow}%
\usepackage{amssymb,amsfonts}%
\usepackage{mathrsfs}%
\usepackage[title]{appendix}%
\usepackage{xcolor}%
\usepackage{textcomp}%
\usepackage{manyfoot}%
\usepackage{booktabs}%
\usepackage{algorithm}%
\usepackage{algorithmicx}%
\usepackage{algpseudocode}%
\usepackage{listings}%
\usepackage{xcolor}
\begin{document}

\title{Building Trustworthy AI: Transparent AI Systems via Language Models, Ontologies, and Logical Reasoning (TranspNet)}
\titlerunning{Building trustworthy AI: TranspNet)}
%
\author{Fadi Al Machot\orcidID{0000-0002-1239-9261}\and Martin Thomas Horsch\orcidID{0000-0002-9464-6739} \and 
Habib Ullah\orcidID{0000-0002-2434-0849}} 
\authorrunning{F.~Al Machot et al.}
%
\institute{Department of Data Science, Faculty of Science and Technology, Norwegian University of Life Sciences, P.O.~Box 5003, 1432 \AA{}s, Norway \\
\email{\{fadi.al.machot, martin.thomas.horsch, habib.ullah\}@nmbu.no}}
\maketitle              

\begin{abstract}
Growing concerns over the lack of transparency in AI, particularly in high-stakes fields like healthcare and finance, drive the need for explainable and trustworthy systems. While Large Language Models (LLMs) perform exceptionally well in generating accurate outputs, their "black box" nature poses significant challenges to transparency and trust. To address this, the paper proposes the TranspNet pipeline, which integrates symbolic AI with LLMs. By leveraging domain expert knowledge, retrieval-augmented generation (RAG), and formal reasoning frameworks like Answer Set Programming (ASP), TranspNet enhances LLM outputs with structured reasoning and verification.This approach strives to help AI systems deliver results that are as accurate, explainable, and trustworthy as possible, aligning with regulatory expectations for transparency and accountability. TranspNet provides a solution for developing AI systems that are reliable and interpretable, making it suitable for real-world applications where trust is critical.
\end{abstract}


\section{Introduction}
Symbolic AI, a fundamental branch of artificial intelligence, focuses on using structured representations of knowledge and formal logic to simulate human reasoning, problem-solving, and decision-making processes \cite{wan2024towards}. Unlike data-driven approaches such as machine learning and Large Language Models (LLMs), which derive patterns from vast amounts of unstructured data, symbolic AI models intelligence is based on  rule-based systems that explicitly define rules, relationships, and logical structures. These structures include formal logic, ontologies, and semantic networks, enabling symbolic AI to work with well-defined rules to draw logical conclusions and make interpretable decisions \cite{dinu2024symbolicai}.

A core component of symbolic AI is knowledge representation, where symbols denote real-world entities and their relationships. Representations in symbolic AI often employ hierarchical or graph-based structures such as semantic networks or ontologies~\cite{baader2017dl,gruber1993translation}. In contrast, LLMs GPT4~\cite{openai2023gpt} and BERT~\cite{devlin2018bert}, while highly effective in generating contextually relevant responses, face challenges in offering the same level of explainability due to their "black box" nature \cite{zhu2024exploring}. This gap is particularly concerning in high-stakes domains like healthcare, finance, and legal reasoning, where trust and transparency are paramount.


This gap between the complexity of LLMs and the demand for transparency poses a significant challenge for AI development, particularly given the legal requirements imposed by regulations like the EU’s General Data Protection Regulation (GDPR), which grants individuals the right to receive an explanation when subjected to automated decision-making processes~\cite{goodman2017european}.

In addition, the proposed AI Act in the European Union, for example, mandates that AI systems—particularly those relying on LLMs in high-risk applications—be designed with human oversight, transparency, and risk management at their core~\cite{aiact2024}. One of the key barriers to achieving trustworthiness with LLMs is the inherent uncertainty in their predictions, which can be influenced by factors such as noisy data, biases in training data, or insufficient training on edge cases. Addressing these uncertainties while maintaining transparency and trustworthiness is a critical challenge in developing LLM-based systems for real-world applications~\cite{peng2011reproducible}.

To address this, systems like the proposed TranspNet pipeline combine the strengths of LLMs with symbolic AI by integrating domain expert knowledge, retrieval-augmented generation (RAG), and formal reasoning frameworks like Answer Set Programming (ASP). This hybrid approach allows LLMs to benefit from the structured, logical reasoning of symbolic AI, ensuring that their outputs are not only accurate but also explainable and trustworthy \cite{machot2023bridging}. Ontologies play a key role in this integration, providing a framework for verifying LLM outputs and enhancing their transparency, a characteristic central to symbolic AI \cite{wan2024towards}.

By combining the power of LLMs with formal reasoning, retrieval-augmented generation, multimodal data processing, and robust documentation practices, our pipeline addresses the key challenges associated with explainability and trustworthiness in AI systems. The integration of formal logic through ASP further distinguishes our approach by providing a mechanism for verifying the logical soundness of the LLM's outputs and addressing uncertainty in a structured and interpretable manner.

\section{State-of-the-Art in LLM Reasoning}

Large language models (LLMs) have revolutionized natural language processing, achieving breakthrough performance on tasks like translation, summarization, and question-answering through in-context learning, a new paradigm that enables few-shot learning without modifying model parameters~\cite{radford2019language,vaswani2017attention,wei2021finetuned}. These models have demonstrated exceptional proficiency in what Kahneman~\cite{daniel2017thinking} describes as ``System 1'' tasks -- automatic, intuitive operations -- but have faced challenges in ``System 2'' reasoning tasks, which require conscious, logical steps, such as solving math word problems~\cite{cobbe2021training}.

Recent developments in prompt-based reasoning, such as Chain-of-Thought (CoT) prompting, have been instrumental in addressing these challenges. By guiding LLMs to generate intermediate reasoning steps, CoT has significantly improved performance on reasoning-intensive benchmarks, such as the GSM8K dataset for math word problems~\cite{wei2021finetuned}. This approach, along with advances in self-consistency and prompt-engineering techniques, has helped bridge the gap between LLMs' ability to perform associative tasks and their capacity for multi-step reasoning~\cite{kojima2022large}.

Despite these advancements, challenges remain in ensuring the faithfulness and interpretability of reasoning processes in LLMs. Techniques like self-verifica\-tion and reinforcement learning have been employed to minimize errors and improve reliability, but issues such as hallucination and error accumulation persist, particularly in complex, multi-step reasoning tasks~\cite{wang2024q}. Additionally, research continues to explore how reasoning capabilities can be transferred to smaller models or embodied agents, as computational efficiency becomes an increasingly important factor in the deployment of LLMs~\cite{meadows2024exploring}.

The proposed pipeline integrates \textit{domain expert knowledge}, \textit{retrieval-augmented generation}, and \textit{formal reasoning frameworks}, enabling the system to verify and refine its outputs through external data and structured reasoning. This multi-layered approach mitigates some of the inherent issues in LLMs, such as hallucination and over-reliance on probabilistic reasoning. The inclusion of ASP~\cite{gelfond1988stable} in the pipeline ensures that the reasoning process is grounded in logical consistency, providing verifiable and interpretable outputs. By incorporating \textit{multimodal data processing}, the pipeline allows the system to handle diverse types of input, further improving its reliability and decision-making capabilities across different domains.

\section{Methodology}

\begin{figure}[h!]
    \centering
    \includegraphics[width=\linewidth]{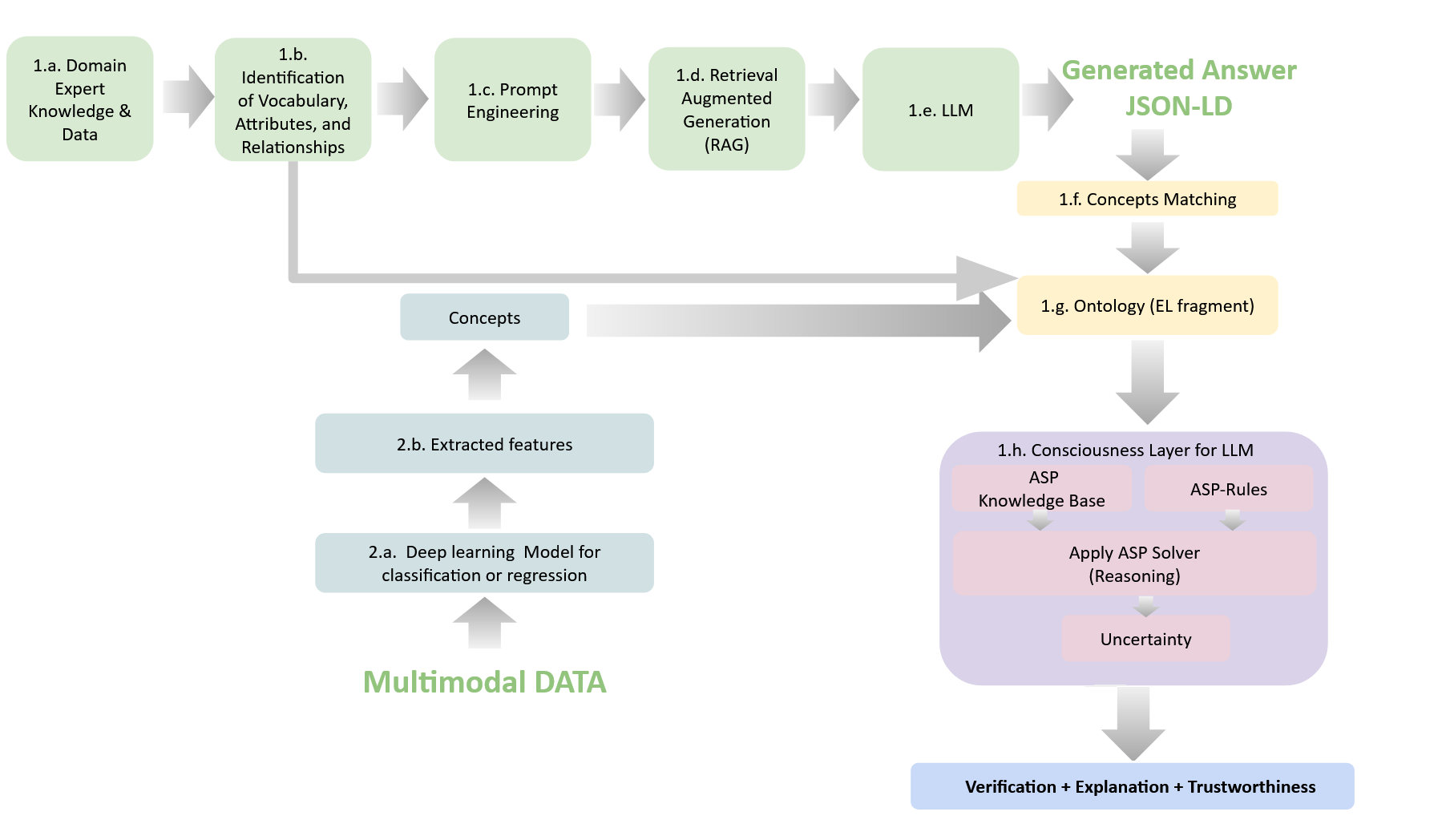}
    \caption{Overall knowledge-driven LLM verification pipeline. The pipeline integrates domain expert knowledge, prompt engineering, retrieval-augmented generation, answer set programming, and deep learning models  to support the generation of accurate, explainable, and trustworthy outputs from LLMs.}
    \label{fig:overall_pipeline}
\end{figure}
\subsection{Overall Knowledge-Driven LLM Verification Pipeline}

The proposed pipeline integrates domain expert knowledge and structured data to enhance the reliability and LLM outputs. The process starts with expert knowledge and vocabulary analysis, followed by prompt engineering and RAG to leverage large language models for generating answers. Ontology matching ensures semantic consistency, with structured ontologies being used for mapping concepts. A critical component of the pipeline is the \textit{consciousness layer for LLMs}, which incorporates ASP for applying logical reasoning. This layer enhances the verification, explanation, and trustworthiness of the generated answers, leading to a robust system aimed at providing accurate and explainable outputs~\cite{baral2003knowledge,erdem2016applications}.

\subsection{Domain Expert Knowledge \& Data (1.a)}

The pipeline begins with the input from domain experts and relevant structured data. The data may include technical documents, research papers, and domain-specific databases.

\subsection{Identification of Vocabulary, Attributes, and Relationships (1.b)}

The identification of vocabulary, attributes, and relationships is a foundational step in the pipeline, ensuring semantic consistency and alignment with domain-specific knowledge. This process involves defining key terms (vocabulary), their properties (attributes), and the logical or causal connections between them (relationships). For example, in healthcare, vocabulary might include terms like "symptom," "diagnosis," and "treatment," with attributes such as severity or duration, and relationships such as "symptom X is linked to diagnosis Y." This step can be performed manually by domain experts, who ensure precision and relevance, or automated using tools like natural language processing and machine learning to extract terms and relationships from structured datasets or research papers. Often, a hybrid approach is used, where automated methods handle initial extraction, and experts refine or validate the results. 

\subsection{Prompt Engineering for Consistent Triple Generation (1.c)}

At this stage, the LLM is prompted to generate responses in a consistent structured format, specifically using (subject-predicate-object) triples in JSON-LD format~\cite{kellogg2019json}. Prompt engineering ensures that the controlled vocabulary is used consistently in the LLM output.

\textbf{Prompt Example:} 
\begin{quote}
``Using the following vocabulary list, generate responses in the form of (subject-predicate-object) triples. Ensure all terms are used consistently according to the provided definitions: [vocabulary list].''
\end{quote}

\subsection{Retrieval-Augmented Generation (1.d)}

RAG combines both retrieval-based and generation-based methods to enhance the output of LLMs. The retrieval model, typically a dense passage retriever (DPR), fetches relevant documents from a large corpus based on the input query. The generator model, usually a transformer, then uses this context to produce coherent and contextually accurate responses. This hybrid approach leverages the strengths of both retrieval and generation techniques, improving the overall relevance and quality of the generated outputs.

\subsection{Large Language Model Generation (1.e)}
Large Language Models (LLMs) are employed for additional tasks such as refining responses, summarization, and generating structured outputs.

In this pipeline, Retrieval-Augmented Generation (RAG) first retrieves relevant context from external sources. This context is then provided to the LLM to guide its response generation, ensuring outputs are grounded in reliable and contextually relevant information.

After the input query is refined through prompt engineering, the LLM generates structured responses in a subject-predicate-object format. Leveraging its extensive training on diverse datasets, the LLM produces contextually accurate outputs aligned with the controlled vocabulary defined in earlier steps. These generated triples are subsequently mapped to the ontology, enabling advanced reasoning and analysis in downstream processes.

\subsection{Concept Matching (1.f)}

Concept matching compares elements in the generated triples with those in the ontology to ensure alignment. Multiple matching techniques are used, including:
\begin{itemize}
     
\item \textbf{Name-Based Matching:} String matching algorithms like Levenshtein distance and Jaccard similarity to compare element names~\cite{euzenat2007ontology}.
\item  \textbf{Structure-Based Matching:} Examination of class hierarchies and property relationships in the schema~\cite{melnik2002similarity}.
\item  \textbf{Instance-Based Matching:} Comparison of actual data instances for value similarity~\cite{doan2001reconciling}.
\item  \textbf{Linguistic Matching:} Use of natural language processing techniques, such as synonym databases and word embeddings, to find semantically similar matches~\cite{zhang2014ontology}.
\end{itemize}

\subsection{Ontology EL-Fragment (1.g)}

Ontologies structure relationships between concepts within a domain, facilitating better data integration and analysis. The pipeline leverages EL fragments (a Description Logic EL fragment)~\cite{xiong2022el}, which supports ASP by enhancing the explainability and tractability of generated data. EL fragments ensure consistent terminology, efficient mapping of triples, and logical reasoning in a scalable manner, making it ideal for complex applications like biomedical informatics.

\subsection{Consciousness Layer for LLM (1.h)}
The role of the Consciousness Layer is to reason on the concepts extracted from (2.a) and the structured response from LLM (1.e). Therefore,  the required piece of information into the domain specific ontology will be use as a knowledge-base for the ASP solver.
The Consciousness Layer includes:
\begin{itemize}
    \item  \textbf{ASP Knowledge Base:} Stores factual and procedural knowledge from mapped concepts and domain knowledge.
    \item \textbf{ASP Rules:} Logical rules and constraints for reasoning over the knowledge base.
    \item  \textbf{ASP Solver:} The ASP solver applies logical inference to refine LLM-generated answers, ensuring consistency, accuracy, and robustness. This solver verifies that the generated triples are logically sound and contextually relevant~\cite{gebser2022answer}.
\end{itemize}

\subsection{Deep Learning Models for Feature Extraction (2.a)}

To address classification, regression, clustering, and time-series problems, deep learning models could be used for feature extraction. These models are capable of extracting rich features that are then mapped to relevant concepts using techniques like DeViL (Decoding Vision features into Language)~\cite{dani2023devil}.

\subsection{Extracted features (2.b)}
Extracted features refer to the characteristics obtained after applying deep learning models for tasks such as classification, regression, or clustering. These models process data from various multimodal sources, including sensors, to extract meaningful and relevant features. This approach is particularly advantageous for our use cases, for example in the Chemical Processes context, where precise feature extraction from multimodal sensor data is crucial for accurate analysis and prediction.
To further enhance the interpretability and utility of these extracted features, it is important to map them to higher-level concepts. Techniques such as the bottleneck model, inspired by the DeVil model, can be employed to identify and distill key features that contribute most significantly to the model’s performance. By identifying these critical features and relating them to specific concepts, we can achieve a more intuitive understanding of the model's decision-making process.

\section{Use-Cases}
\subsection{Use Case 1: Healthcare - Clinical Decision Support System}

In healthcare, clinical decision support systems (CDSS) are used to assist physicians in making informed treatment decisions based on patient data and medical literature. The proposed pipeline can enhance these systems by ensuring the reliability, explainability, and trustworthiness of the LLM outputs used in patient care. The pipeline begins by integrating medical knowledge from domain experts, including data from medical research papers, clinical guidelines, patient history, and real-time data from electronic health records (EHR). This step ensures that the LLM receives accurate and up-to-date medical knowledge relevant to clinical decision-making.

In the next step, key medical terms, conditions, symptoms, diagnostic tests, and treatment options are identified and structured. This phase ensures that the LLM understands the relationships between diseases, symptoms, and treatments, facilitating more accurate clinical recommendations. Using the identified vocabulary, the LLM is prompted to generate responses related to patient conditions, diagnostics, and treatments in a structured format like (patient-symptom-disease). The system then retrieves relevant medical literature and studies using RAG to back up recommendations, ensuring that the generated outputs are grounded in evidence.

The LLM generates structured outputs, such as (patient has symptoms X, Y, Z - potential diagnosis: Disease A), based on patient data and retrieved medical information. Medical ontologies, such as SNOMED CT~\cite{chang2021use}  or ICD-10~\cite{hussey2022snomed}, are used to match the generated triples, ensuring semantic consistency and accuracy in the medical domain. The ASP-based consciousness layer ensures that the recommendations made by the LLM are logically sound and aligned with clinical guidelines, reducing the risk of incorrect recommendations. This pipeline ensures that the clinical decision support system produces trusted and explainable medical recommendations for healthcare providers, with outputs that are clinically accurate, evidence-based, and compliant with medical guidelines.

\subsection{Use Case 2: Battery Design - Material Selection for Energy Storage}

In battery design, selecting the right materials for components such as electrodes and electrolytes is critical to improving battery efficiency and lifespan. The proposed pipeline supports engineers and scientists by providing reliable and explainable recommendations on material combinations for energy storage systems. The pipeline starts by integrating data from scientific literature, material databases, and experimental results related to the battery technology. This input includes expert knowledge on chemical properties, performance metrics, and degradation, which guides material selection.

The next step involves identifying key attributes such as material conductivity, chemical stability, energy density, and thermal properties. Relationships between these attributes and battery performance (e.g., how conductivity affects charge/discharge rates) are mapped to ensure the LLM understands the complexity of material behavior in battery systems. The LLM is then prompted to generate suggestions for materials based on performance criteria, such as (material X - conductivity Y - potential application Z). For example, the LLM might propose materials with high conductivity and thermal stability for use in lithium-ion battery electrodes.

Next, RAG retrieves recent research papers and experimental data on materials from scientific databases. If the LLM suggests graphene as a suitable material for a battery component, RAG pulls relevant studies on graphene’s performance in energy storage applications to substantiate the recommendation. The LLM generates structured outputs suggesting material combinations in (material-property-application) triples. For instance, it might recommend using graphene for electrodes due to its high conductivity and chemical stability. Material ontologies and databases are used to match the generated triples, ensuring alignment with the known properties of materials.

Finally, the ASP-based Consciousness Layer applies logical rules to verify that the suggested materials meet the specific requirements of the battery design. For example, if a material is recommended for high-temperature environments, ASP checks whether the material's thermal stability is sufficient, ensuring logical consistency and trustworthiness. The pipeline produces explainable, evidence-based recommendations for material selection in battery design, helping engineers and scientists optimize material choices for performance and longevity, with clear explanations of the decision-making process.

\section{Limitations}
One limitation of the proposed pipeline is that its emphasis on structured reasoning and logical consistency might reduce the flexibility and adaptability of LLMs in certain contexts. For example, in applications like healthcare or education, the pipeline excels by ensuring outputs are accurate, reliable, and aligned with established guidelines. This is especially important when making decisions that directly impact patient safety or the clarity of learning materials.

However, this structured approach might not be as effective in more creative or exploratory applications, such as marketing, content creation, or brainstorming new product ideas. In these scenarios, flexibility, innovation, and the ability to generate unconventional or imaginative responses are often more valuable than strict logical accuracy. By limiting the LLM's predictive power to fit within predefined rules and structures, the pipeline may constrain its potential to explore diverse or unexpected outcomes.

\section{Conclusion}
In this work, we presented a pipeline designed to enhance the explainability, accuracy, and trustworthiness of LLMs. By integrating domain expert knowledge, prompt engineering, RAG, and ASP, the proposed pipeline addresses key challenges in ensuring that LLM-generated outputs are logically consistent, contextually relevant, and semantically aligned with domain-specific knowledge. 

The proposed pipeline represents an advancement in addressing the inherent limitations of traditional LLMs. Its ASP-based Consciousness Layer sets a  benchmark for ensuring logical consistency, enhancing trust, and fostering greater transparency in AI outputs. As LLMs continue to evolve and find applications in increasingly critical domains, approaches that combine structured reasoning and expert knowledge, as exemplified by this pipeline, will be essential for reliability, transparency, and success in real-world implementations.

\begin{credits}
\subsubsection{\ackname} Funding is acknowledged from the European Union's Horizon Europe research and innovation programme under grant agreements no.~101137725 (BatCAT) and 101138510 (DigiPass CSA).

\subsubsection{\discintname}
The authors have submitted a grant proposal in which the presented architecture plays a central role.
\end{credits}

%
%
\bibliographystyle{splncs04}
\bibliography{ref}

\end{document}